\documentclass{article}
\usepackage{deari}

\usepackage{times}
\usepackage{soul}
\usepackage{url}
\usepackage[hidelinks]{hyperref}
\usepackage[utf8]{inputenc}
\usepackage[small]{caption}
\usepackage{graphicx}
\usepackage{amsmath}
\usepackage{amsthm}
\usepackage{booktabs}
\usepackage{algorithm}
\usepackage{algorithmic}
\usepackage[switch]{lineno}
\usepackage{amsfonts}
\usepackage{subfigure}
\usepackage{enumitem}

\usepackage{tikz}
\newcommand{\circo}{~\raisebox{1pt}{\tikz \draw[line width=0.6pt] circle(1.1pt);}~}

\soulregister\cite7 
\soulregister\citep7 
\soulregister\citet7 
\soulregister\ref7 
\soulregister\pageref7 

\DeclareUnicodeCharacter{2061}{}

\title{Uncertainty-Aware Deep Attention Recurrent Neural Network for Heterogeneous Time Series Imputation}

\author{
Linglong Qian$^1$
\and
Zina Ibrahim$^1$\and
Richard JB Dobson$^{1,2,3}$
\affiliations
$^1$Institute of Psychiatry, Psychology and Neuroscience, King’s College London\\
$^2$University College London\\
$^3$Health Data Research UK, University College London\\
\emails
\{linglong.qian, zina.ibrahim, richard.j.dobson\}@kcl.ac.uk
}

\begin{document}

\maketitle

\begin{abstract}
Missingness is ubiquitous in multivariate time series and poses an obstacle to reliable downstream analysis. Although recurrent network imputation achieved the state-of-the-art, existing models do not scale to deep architectures that can potentially alleviate issues arising in complex data. Moreover, imputation carries the risk of biased estimations of the ground truth. Yet, one’s confidence in the imputed values is almost always unmeasured or computed post-hoc from model output. We propose DEep Attention Recurrent Imputation (DEARI), which jointly estimates missing values and their associated uncertainty in heterogeneous multivariate time series. By jointly representing feature-wise correlations and temporal dynamics, we adopt a self-attention mechanism, along with an effective residual component, to achieve a deep recurrent neural network with good imputation performance and stable convergence. We also leverage self-supervised metric learning to boost performance by optimizing sample similarity. Finally, we transform DEARI into a Bayesian neural network through a novel Bayesian marginalization strategy to produce stochastic DEARI, which outperforms its deterministic equivalent. Experiments show that DEARI surpasses the state-of-the-art in diverse imputation tasks using real-world datasets, namely air quality control, healthcare and traffic.

\end{abstract}

\section{Introduction}
In many application domains such as finance \cite{finance}, healthcare \cite{healthcare} and weather prediction \cite{schultz2021can}, large and heterogeneous multivariate time series constitute the majority of data used to perform predictive tasks. However, designing successful classification and regression applications in the corresponding areas requires tackling the abundance of non-randomness missingness in most domains. For example in healthcare, electronic health records (EHRs) are rich resources used to design patient-specific prognostic models \cite{kdop,farah}. However, EHR time-series are generated as a byproduct of patient care and reflect clinical decisions, rendering them naturally irregularly-sampled and plagued with non-random missingness \cite{lipton2016directly}. 

Different flavours of neural networks have achieved success in imputing large and heterogeneous multivariate time-series \cite{gu2018recent,schuster1997bidirectional,arber1997mlp}. The current state-of-the-art, BRITS \cite{cao2018brits} uses a single-layer bidirectional RNN to outperform prior models. Similarly to other highly-performing imputation models, however, BRITS is deterministic, making it difficult to quantify one's confidence in the resulting imputation.

Our work is concerned with filling the current gaps in handling missingness in large multi-variate time-series using BRITS as a backbone. We re-frame the imputation problem as a scalable self-supervised learning problem to alleviate issue of data heterogeneity and devise an uncertainty-aware stochastic learning model. Our aim is to release the fundamental issue of imputation from task-specific biases while achieving high imputation performance through scalable architectures. Our novel DEep Attention Recurrent Imputation (DEARI) model makes the following technical contributions: 

\begin{itemize}[leftmargin=*]
\item We enable deeper architectures by incorporating a self-attention mechanism and a residual component into the backbone BRITS. Our reformulation, which collapses into a BRITS model when a single layer, is highly scalable in-depth, enabling more flexible and generalisable architectures for large and complex datasets.

\item DEARI achieves more accurate imputation by regarding imputation as a self-supervised learning task while simultaneously aiming to learn a good data representation. In heterogeneous data (where sample similarity/dissimilarity may not consistently correlate with class labels) with complex features, forward and backward representations of a single sample from an RNN maybe more homogeneous than any other samples. DEAR implements a deep metric learning (DML) framework on triplets constructed using the two directions for better estimations. 

\item DEARI is equipped with a Bayesian marginalization strategy combining its reliable imputation with trustworthy confidence bounds. We achieve this by incorporating the framework into a Bayesian neural network, exploring its performance under real complex scenarios. DEARI overcomes difficulties in training Bayesian NNs via a simple but practical training strategy to trade off exploration with exploitation, achieving further improvement in model performance over deterministic DEARI in large datasets. 
\end{itemize}
We experimentally evaluate DEARI to demonstrate its capabilities when imputation is released from biases specific to the downstream classification/regression task using five real-world datasets. Our results show that both deterministic and stochastic implementations of DEARI outperform the state-of-the-art models for imputation accuracies.

\section{Related Work}
The literature contains multiple approaches to multivariate time series imputation. We focus on highly-performing models that have been tied with non-random missingness when data is big. The RNN model GRUD \cite{che2018recurrent} introduced the idea of temporal decay \cite{vodovotz2013systems}, i.e. the contribution of past recordings to missingness decreases with time, to impute features closer to default values if their last observation occurred long time ago.

MRNN \cite{yoon2017multi} and BRITS \cite{cao2018brits} extend GRUD to model missingness patterns across features in addition to capturing temporal dynamics and use bidirectional recurrent dynamics to enhance accuracy by estimating missing values in the forward and backward directions. However, imputed values are treated as constants in MRNN and cannot be sufficiently updated. In contrast, BRITS does away with data-specific assumptions achieving the state-of-the-art in several domains. Nevertheless, BRITS, functions as a variant of a single LSTM cell, which is insufficient when dealing with complex missingness or highly-dimensional datasets with significantly high missingness rates. Complex features are known to be better represented by deeper architectures \cite{bengio2009learning,zilly2017recurrent}. We believe that a more scalable architecture can also capitalise on approaches such as deep metric learning \cite{kaya2019deep}, which optimises sample-wise similarity, therefore enabling efficient unsupervised and self-supervised training in complex data and has shown success under multimodal settings \cite{suo2019metric}. 

An important issue directly relates to the ability to quantify confidence in the generated data. Stochasticity has been introduced in RNN models in GRUU \cite{jun2020uncertainty}, variational autoencoders in V-RIN \cite{mulyadi2021uncertainty} and generative adversarial networks in $E^2$GAN \cite{luo2018multivariate}. However, all existing approaches require additional uncertainty modules added to the imputing network, which translates to unstable training due to increased coupling and leads to less accurate imputation\cite{mescheder2018training}. Although $E^2$GAN asserts to resolve the issues of stochastic generative DL imputation models \cite{wang2016auto,kingma2013auto} struggling to model good loss functions, this remains highly untested, and performance against BRITS is unknown. Other stochastic approaches based on GAN \cite{cao2022survey} suffer from unstable training and convergence difficulty \cite{mescheder2018training}.

Our final examination relates to whether imputation is performed jointly with the downstream task (classification/regression) using one neural graph. This approach is adopted by GRUU, V-RIN and BRITS. However, our examination showed that GRUU only improved performance in the predictive task, and V-RIN failed to reproduce good performance. Our position is that the downstream task reduces diversity; the coarse-grained classification task may focus on different parts of the data than the fine-grained imputation task, resulting in model bias when performed simultaneously. To this end, we focus only on imputation performance.

\section{Terminology and Background}

\begin{figure}[h]
\centering
\includegraphics[width=\linewidth]{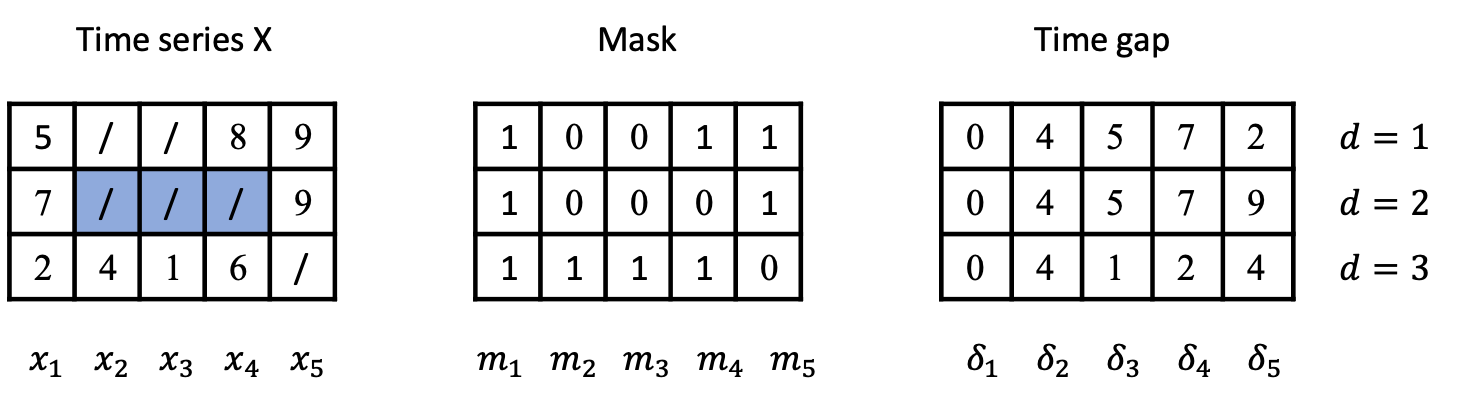}
\caption{An example of multivariate time-series. $x_{1-5}$: observations in time steps $t_1,\,...,\,t_5$ with corresponding time-stamps $s_{1-5}=0,\,4,\,5,\,7,\,9$. Feature $d_2$ was missing during $t_{2-4}$, the last observation took place at $s_1$. Hence, $\delta_5^2 =t_5-t_1 =9-0=9$.}
\label{fig:1}
\end{figure}

We adopt the terminology used by our backbone model, BRITS. For a temporal interval observed over $T$ time-steps. A multivariate time series is denoted by a matrix $\boldsymbol{X}=\{\boldsymbol{x_{t_1},\,x_{t_2},\,...,\,x_{t_T}}\}$ consisting of $T$ observations. Each observation $\boldsymbol{x_{t_n}} \in \mathbb{R}^D$ is a vector of $D$ features $\{x_{t_n}^{1},\,x_{t_n}^{2},\,...,\,x_{t_n}^{D}\}$. Notably different to BRITS, $\mathbb{R}^D$ is not necessarily homogeneous. It may include variables that are numerical, categorical, static or dynamic as well as other structured modalities $\{D_{num},\,D_{cat},\,D_{sta},\,D_{dyn}\}\in \mathbb{R}^D$. Moreover, because the time-gap between two consecutive time-steps may vary, we use $\delta_{t_1},\,...,\,\delta_{t_T}$ to denote the time gaps at time steps $t_1,\,...,\,t_T$ consecutively. Information about missingness is represented by two matrices derived from $\boldsymbol{X}$ (Fig. \ref{fig:1}). The mask $\boldsymbol{M} \in \mathbb{R}^{N \times D}$ describes whether the elements of $\boldsymbol{X}$ are present or missing:

\begin{equation*}
{m_{t_n}^d} = \begin{cases}
{0,}&{\text{if}}\ {x_{t_n}^d \text{ is not observed}} \\ 
{1,}&{\text{otherwise.}} 
\end{cases} 
\end{equation*}

Because the data in $\boldsymbol{X}$ may not be evenly sampled, the time gap between consecutive observations varies across time-stamps. $\delta \in \mathbb{R}^{N \times D}$ provides an indication of sparseness of a feature by encoding information about the time-gap between two consecutive observed cells for each feature as follows:

\begin{equation*}
{\delta_{t_n}^d} = \begin{cases}
{t_n - t_{n-1} + \delta_{t_{n-1}}^d} & {\text{if}}\ {t_n} > 1, {m_{t_n}^d} = 0 \\ 
{t_n - t_{n-1}}  & {\text{if}}\ {t_n} > 1, {m_{t_n}^d} =1 \\ 
{0}   &{\text{if}}\ {t_n} = 0
\end{cases} 
\end{equation*}

\subsection{Overview of the BRITS Backbone}\label{brits}

We build on the BRITS assumptions which state that $\boldsymbol{X}$ contains correlations across the temporal and feature dimensions. The BRITS architecture is composed of a fully-connected regression component and a recurrent component. The temporal correlations are governed by the notion of temporal decay - Eq. \eqref{eq:decay}, which dictates that the strengths of the correlations are inversely related to the time-gap $\delta_t$. The calculated decay factor $\gamma_{th} \in (0,1]$ for the hidden state is then used to transform $h_{t-1}$ to a \emph{decayed} hidden state, $\hat{h}_{t-1} $ as in Eq. \eqref{eq:decayedstate}. 

Because an observation $\boldsymbol{x_t}$ may contain missing values, it cannot be used directly.  BRITS first uses the decayed hidden state to find the historic representation, $\boldsymbol{\hat{x}_t} $ as in Eq. \ref{eq:regression}. Using masking vector $\boldsymbol{m_t}$ obtained from $\boldsymbol{M}$,  $\boldsymbol{\hat{x}_t}$ is then used to replace the missing values of $\boldsymbol{x_t}$, yielding the complement vector $\boldsymbol{x_t^{hc}} $, which embeds missingness patterns to be fed into subsequent BRITS components (Eq. \eqref{eq:complement}).

\begin{align}
\label{eq:decay} \gamma_{th}  & = \exp⁡\{-\max(0,\;\boldsymbol{W_{\gamma h}} \delta_t + \boldsymbol{b_{\gamma h}} )\}  \\
\label{eq:decayedstate} \boldsymbol{\hat{h}_{t-1}} & =  \boldsymbol{h_{t-1}} \odot \gamma_{th} \\
\label{eq:regression} \boldsymbol{\hat{x}_t} & =  \boldsymbol{W_x}\hat{h}_{t-1} + \boldsymbol{b_x}\\
\label{eq:complement} \boldsymbol{x_t^{hc}} & =  \boldsymbol{m_t }\odot \boldsymbol{x_t} + (1-\boldsymbol{m_t}) \odot \boldsymbol{\hat{x}_t}
\end{align}

BRITS further investigates correlations within a single observation. Here, a fully-connected layer composed of the derived $\boldsymbol{x_t^{hc}} $ is used while restricting the diagonal of the parameter matrix $\boldsymbol{W_z}$ to be zeros. Thus, we can obtain $\boldsymbol{x_t^{fc}}$ as the feature-wise estimation of the missing values (Eq. \ref{eq:full}), each $d^{th}$ element is based solely on the other features in $t^{th}$ step.

\begin{equation}
\label{eq:full} \boldsymbol{x_t^{fc}} =  \boldsymbol{W_z x_t^{hc}}+\boldsymbol{b_z}
\end{equation}

BRITS applies the notion of decay to the cross-sectional feature space (Eq. \eqref{eq:decayspatial} to obtain a learnable factor $\hat{\beta}_{t} $ (Eq. \eqref{eq:beta}), which considers both the temporal decay $\gamma_{tf}$ for features and the masking vector $\boldsymbol{m_t}$, therefore combining temporal and feature-wise estimations of missingness patterns.

\begin{align}
\label{eq:decayspatial} \gamma_{tf}  & = exp⁡\{-(0,\;\boldsymbol{W_{\gamma f} }\delta_t +\boldsymbol{b_{\gamma f}})\} \\
\label{eq:beta} \hat{\beta}_{t} & = \sigma(\boldsymbol{W_\beta} [\gamma_{tf} \circo \boldsymbol{m_t}] +\boldsymbol{b_\beta})
\end{align}

Using $\hat{\beta}_{t}$ and the same notion of complement (Eq. \eqref{eq:compbeta}) will result in the replacements of the missing values in the original data, generating  the imputation $\boldsymbol{C_t}$ (Eq. \eqref{eq:imputed}.

\begin{align}
\label{eq:compbeta} \boldsymbol{x_t^c} & =   \beta_t \odot \boldsymbol{x_t^{fc}} + (1-\beta_t) \odot \boldsymbol{x_t^{hc}}\\
\label{eq:imputed} \boldsymbol{C_t} & =  \boldsymbol{m_t }\odot \boldsymbol{x_t} + (1-\boldsymbol{m_t}) \odot \boldsymbol{x_t^c}
\end{align}

The final step (Eq. \eqref{eq:rnn}) is to update the hidden state, where RNNs use indicators to learn arbitrary functions of past observations and missingness patterns. Indicators can be any informative factor, both GRUD, BRITS and V-RIN using imputed values and corresponding masks.

\begin{equation} \label{eq:rnn}
   \displaystyle \boldsymbol{h_t} = \sigma(\boldsymbol{W_t \hat{h}_{t-1}} + \boldsymbol{U_h }[ \boldsymbol{C_t }\circo \boldsymbol{m_t}] + \boldsymbol{b_h})
\end{equation}

Bidirectional recurrent dynamics, used to overcome slow convergence and model backward information, are achieved by reversing the backward imputation and compounding with forward imputation (Eq. \eqref{eq:bidirectional}). $\{\boldsymbol{C_{t_F}},\;\boldsymbol{C_{t_B}}\},\;\{\boldsymbol{h_{t_F}},\;\boldsymbol{h_{t_B}}\}$ are the results of pair-wise outputs.

\begin{equation} \label{eq:bidirectional}
    \boldsymbol{C_t^*} = \frac{\boldsymbol{C_{t_F}}+\boldsymbol{C_{t_B}^T}}{2}
\end{equation}

\section{Methodology}

\begin{figure*}[h]
\centering
\subfigure[DEARI]{
\begin{minipage}[t]{0.23\linewidth}
\centering
\includegraphics[width=\linewidth]{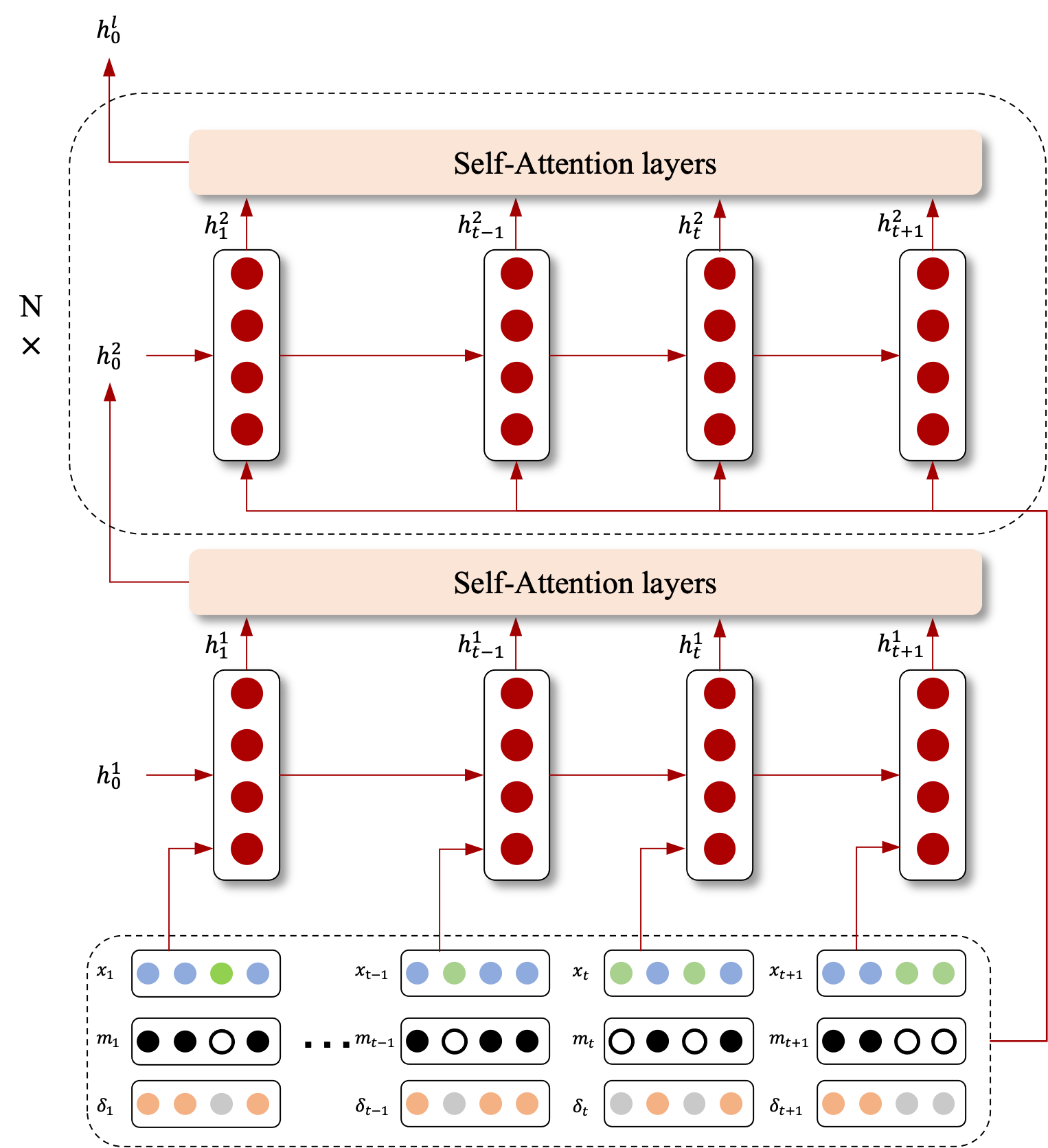} 
\end{minipage}
}
\subfigure[Deep Metric Learning]{
\begin{minipage}[t]{0.3\linewidth}
\centering
\includegraphics[width=\linewidth]{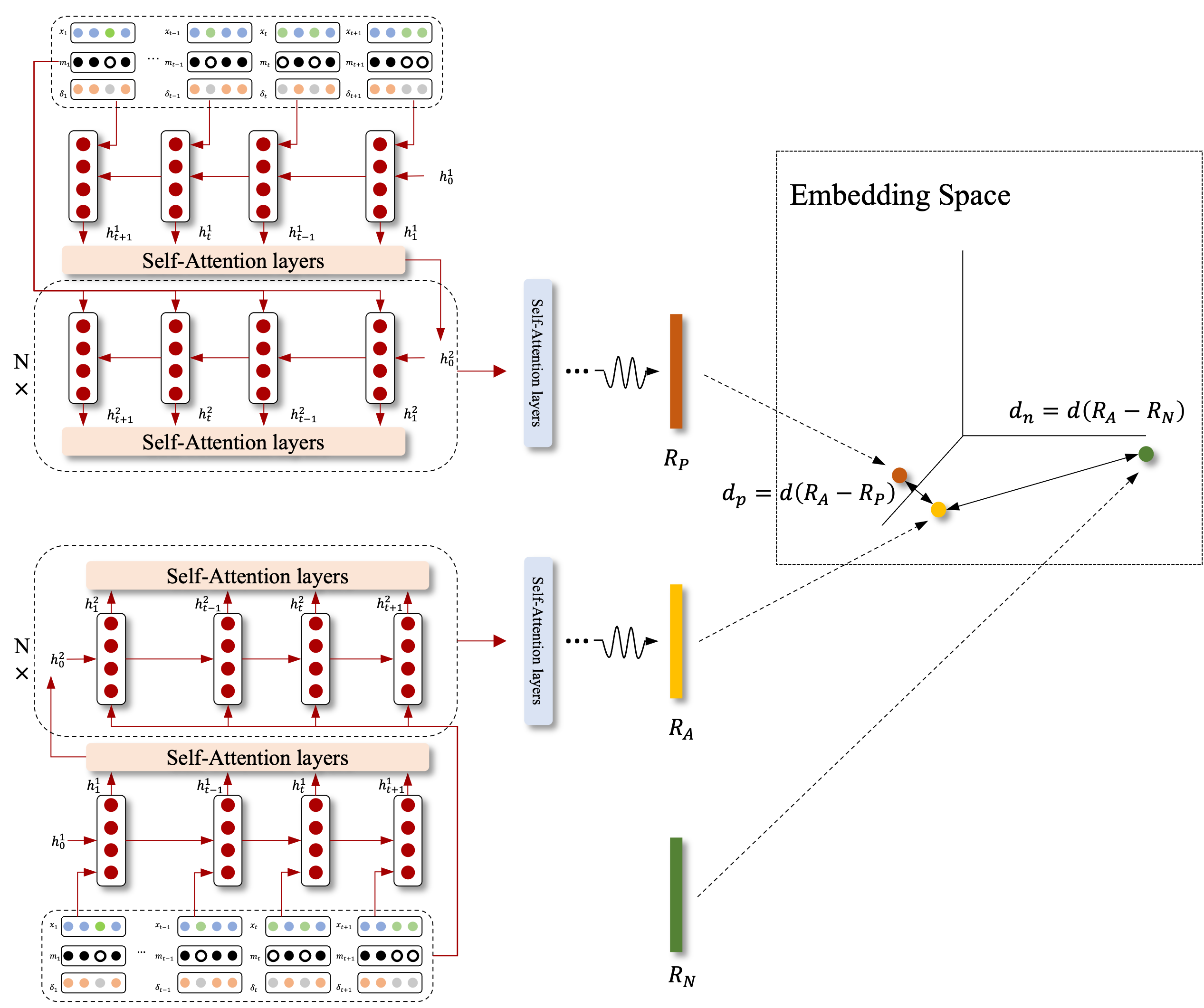}
\end{minipage}
}
\subfigure[Bayesian]{
\begin{minipage}[t]{0.37\linewidth}
\centering
\includegraphics[width=\linewidth]{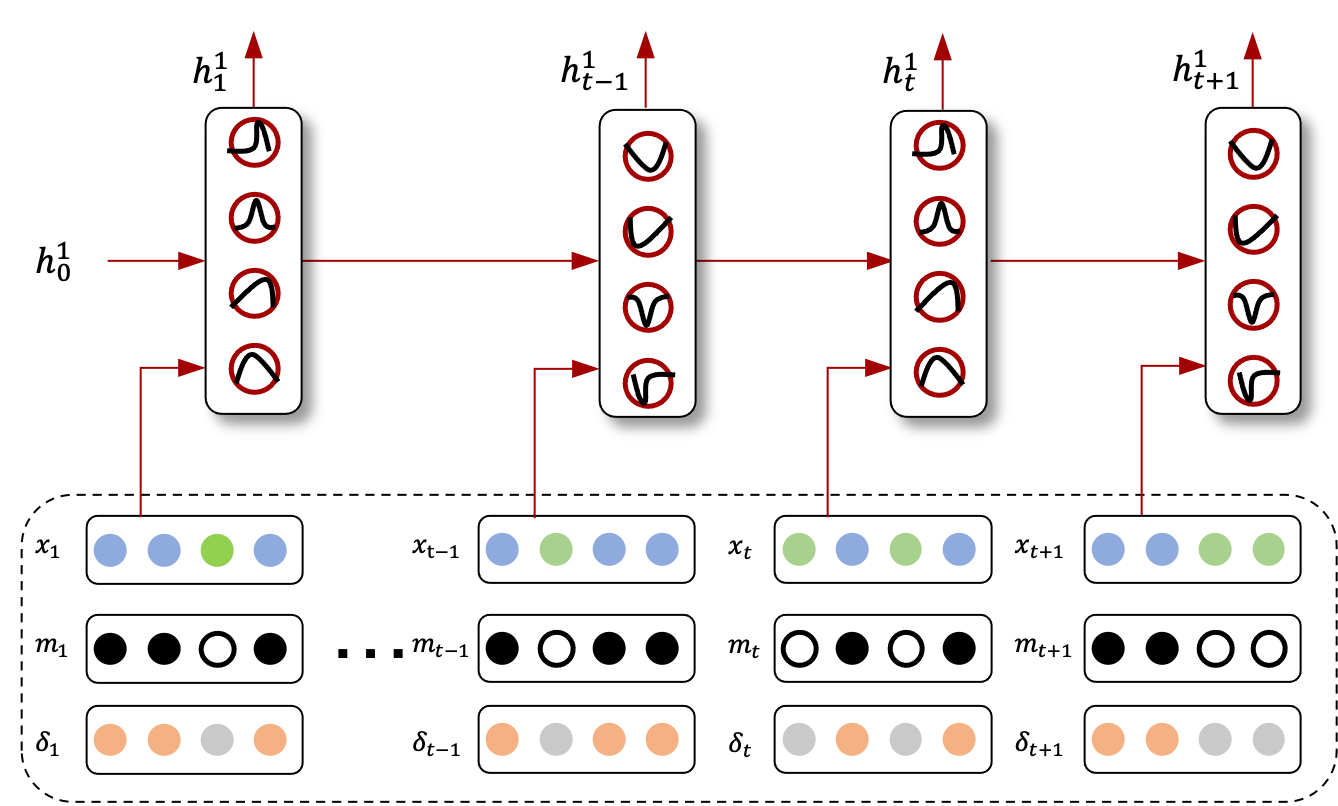}
\end{minipage}
}
\caption{The structure of the proposed methods. a) deep-attention recurrent neural network, which collapses to BRITS if No. layers = 1. b) deep self-supervised metric learning, larger batch sizes lead to more triples. c) Bayesian marginalization, all parameters of the Bayesian layers are controlled by distributions.}
\label{fig:models}
\end{figure*}

We now present the three components that enable scalable and flexible imputation via DEARI (Fig. \ref{fig:models}), namely: 1) \textbf{the deep-attention recurrent neural network}, which is modified multilayer BRITS that has been augmented by self-attention mechanisms, 2) \textbf{the deep self-supervised metric learning}, compounded with self-attention as well. 3) \textbf{Bayesian stochastic marginalization}, which quantifies uncertainty using Monte Carlo simulations.

\subsection{Deep Attention Recurrent neural network}

When multiple RNN cells are stacked, the hidden state of a layer is usually the input for the next layer:

\begin{align}
\boldsymbol{h_t}  & = \sigma(\boldsymbol{W_x x_t} + \boldsymbol{W_h h_{t-1}} + \boldsymbol{b}) \\
\boldsymbol{h_t^l} & = \sigma(\boldsymbol{W_x' h_t^{l-1}} + \boldsymbol{W_h' h_{t-1}^{l}} + \boldsymbol{b'})
\end{align}

BRITS receives incomplete data with missing indicators for imputation. In contrast to conventional RNNs, it is more concerned with accurate imputation than with good embedding. However, continuous representation (hidden state) learning by simple stacking without specific design would amplify the error caused by missingness.

Furthermore, while generating new imputations, BRITS replaces missingness at the start with the decayed hidden state and focuses on learning missing patterns from the observations and corresponding indicators. The stacked layers would ignore the estimated missing values, rendering them unable to contribute (For more information, see Sec~\ref{brits}).

Therefore, we provide our deep layers with both the raw data and the output of the shallow layer. Within one process, the hidden state $h_{t-1}$ mainly effect through decayed state \eqref{eq:decayedstate}, historical estimation \eqref{eq:regression} and hidden state update \eqref{eq:rnn}. Through comparative experiments\footnote{The optimal DEARI is determined by identifying the aggregated information and locations (See details in the supplementary file)}, We found that performance improvement is most noticeable when the initialization of the current hidden state interacts directly with the previous hidden states.

\begin{equation} \label{eq:initialization}
   \displaystyle \boldsymbol{h_0^l} = f([\boldsymbol{h_{0}^{l-1}},\;\boldsymbol{h_{1}^{l-1}},\;...,\;\boldsymbol{h_{t}^{l-1}}])
\end{equation}

\paragraph{Self-Attention Initialization} We further explore the fusion of hidden states. The ordinary RNN is deep with respect to time, with each data stream learned sequentially without any leakage of prior information. According to other neural networks, low-level and high-level information from RNNs also exists, albeit slightly differently. Short-term memory (early time steps) stores information with limited capacity and high density, while long-term memory (later time steps) focuses on information over the whole period and makes it easy to forget details from the early time steps. Even the RNN itself is insufficient to handle long-distance dependencies. Consequently, the last hidden state is typically believed to include information about the entire time series and is regarded as the new representation.

Inspired by the Feature Pyramid Network in computer vision tasks \cite{lin2017feature}, all hidden state columns concatenated together can be considered as a temporal pyramid map at the same column-wise scale. To improve the performance of the fine-grained imputation task, it is necessary to combine the basic semantic feature columns  with the rich semantic feature columns. Therefore, we adopt the self-attention mechanism \cite{vaswani2017attention} to abstract better embedding of the temporal pyramid map.

Similarity to BERT’s [class] token \cite{47751}, we prepend a learnable embedding to hidden states. The transformer encoder consists of alternating layers of multiheaded self-attention (MSA) and feed-forward networks (FFN), along with residual connections and layernorm (LN). $\boldsymbol{h_{0}^{l}}$ is then generated by being processed\footnote{For the first layer, initialize the $\boldsymbol{h_{0}^1}$ by all zero.} as follows:

\begin{align}
\label{eq:cls}   \boldsymbol{\hat{h}^{l-1}} & = [\text{CLS},\;\boldsymbol{h_{0}^{l-1}},\;\boldsymbol{h_{1}^{l-1}},\;...,\;\boldsymbol{h_{t}^{l-1}}] \\
\label{eq:mas}   \boldsymbol{(\hat{h}^{{l-1}})^{\prime}} & = \text{LN}(\text{MSA}(\boldsymbol{\hat{h}^{l-1}}) + \boldsymbol{\hat{h}^{l-1}}) \\
\label{eq:ffn}   \boldsymbol{(\hat{h}^{{l-1}})^{*}} & = \text{LN}(\boldsymbol{(\hat{h}^{{l-1}})^{\prime}}) + \boldsymbol{(\hat{h}^{{l-1}})^{\prime}}) \\
                 \boldsymbol{h_{0}^{l}} & =  \text{CLS}^{*} \in \boldsymbol{(\hat{h}^{{l-1}})^{*}}
\end{align}

However, training deep RNNs (more than 5 layers) remains a challenge, even when LSTM gating mechanisms are used\cite{turkoglu2021gating}. To this end, another special design of our model is that each deep layer can access the raw data $\boldsymbol{X}$ as the residual component\footnote{The mean pooling of all layer losses served as the overall loss.}. The total \textbf{DEARI} guarantees that the model achieves good performance and reliable convergence, even when scaled to 10 layers.

\subsection{Deep Self-Supervised Metric Learning}

In real-world temporal data, different domains show varied heterogeneity. Healthcare data, for instance, shows considerable class imbalance. The time-varying statistics of different types of urban data are challenging to comprehend for a single location, and even more complicated for multiple locations. Making it impractical to incorporate all relevant prior knowledge into the analytical model. Traditional metric learning \cite{kulis2013metric,yang2006distance} automatically constructs task-specific distance metrics from data. On top of the basic model, we apply self-supervised deep metric learning (S$^{2}$DML) to eliminate task bias while exploring individual properties. 

\paragraph{Online Hard Triplets Mining} The directionality of representations from single-layer BRITS or our DEARI is utilized to guide the construction of triplets. While one serves as the anchor $\boldsymbol{R_{A}}$ and another as the positive sample $\boldsymbol{R_{P}}$, the representations from other the samples within the same mini-batch are all negative samples $\boldsymbol{R_{N}}$. Identically to Eq. \eqref{eq:cls} $\sim$ Eq. \eqref{eq:ffn}, we adopt the self-attention mechanism on the final hidden states to obtain the best representation of each direction:

\begin{align} \label{eq:dml_attention}
   \boldsymbol{R_{A},R_{P}} \in & \; [f_{r}(f_{SA}(\boldsymbol{H}_{nF})),\;f_{r}(f_{SA}(\boldsymbol{H}_{nB}))] \\
   \boldsymbol{R_{N}} \in & \; [f_{r}(f_{SA}(\boldsymbol{H}_{mF})),\;f_{r}f_{SA}(\boldsymbol{H}_{mB}))], m \neq n \nonumber
\end{align}
where the $\boldsymbol{H}_{nF}$ represents the forward hidden states of sample $n$, $\boldsymbol{H}_{nB}$ for backward. $f_{r}(*)$ can be one of the \textbf{CLS}-token, last-token and the mean pooling strategies to compute the final representation. Among the constructed triplets, we select out all triplets that violate the following condition:

\begin{equation} \label{eq:triplets}
    \parallel \boldsymbol{R_{A}}-\boldsymbol{R_{P}}\parallel _{2} + \lambda \; < \; \parallel \boldsymbol{R_{A}}-\boldsymbol{R_{N}}\parallel _{2}
\end{equation}
$\lambda$ is a pre-set margin, meaning we only consider samples that are easily confused with the anchor sample by a margin $\lambda$.

\paragraph{Multilayer Adaptation} One of the differences between BRITS and DEARI is that we have multiple outputs. Squeezing them into a single pair (same as the single-layer BRITS) works better than increasing the number of triplets.

\paragraph{Loss Function} A SOTA metric learning objective Multi-Similarity loss \cite{wang2019multi} is used in this work. The MS loss reweighs the importance of the samples by exploiting similarities among and between positive and negative pairs. The more informative ones represent indistinguishable pairs and will therefore contribute more during training.

\begin{align} \label{eq:ms_loss}
    \boldsymbol{\mathcal{L}} = \frac{1}{\mid\mathcal{B}\mid} \sum_{i \in \mathcal{B}}\Bigg\{&\frac{1}{\alpha}\log[1+\sum_{n\in \mathcal{N}_{i}}e^{\alpha(\boldsymbol{\rm{S}}_{in}-\epsilon)}] \\
    +&\frac{1}{\beta}\log[1+\sum_{p\in \mathcal{P}_{i}}e^{\beta(\boldsymbol{\rm{S}}_{ip}-\epsilon)}] \Bigg\} \nonumber
\end{align}
$\alpha, \beta$, $\epsilon$ are hyperparameters; $\mathcal{B}$ denotes mini-batch samples.

\subsection{Bayesian Marginalization Strategy}

It is desirable to have precise predictions, but it would be much better to provide corresponding trustworthy bounds. In this work, we propose the Bayesian Marginalization Strategy \cite{mackay1992practical} to represent the epistemic uncertainty from modelling error or lack of perfect knowledge about the black box model \cite{neal2012bayesian}. Simultaneously, the aleatoric uncertainty from inherent noise perturbation embedded in the data itself is captured by the random masking approach.

General NNs focus on finding one optimal set of weights (optimization), while BNNs interested in the distributions behind weights (marginalization) \cite{dawid1973marginalization}. Uncertainty is then explicitly represented by computing the variability of weights. Having a distribution instead of a single value also makes it possible to evaluate the robustness of the model. In other words, estimating with certainty is a separate task from developing confidence estimates.

Estimating a posterior distribution of weights is practically intractable since it is equivalent to using an ensemble of an infinite number of neural networks \cite{blundell2015weight}. Variational Inference (VI) \cite{blei2017variational} transform the posterior estimation problem into an optimization problem by minimizing the Kullback-Leibler (KL) divergence between the approximate distribution with the truly Bayesian posterior. 

Missingness patterns are mainly observed by the hidden states and updated iteratively. We therefore transform our BRITS backbone into a Bayesian recurrent neural network by replacing each parameter of the RNN components by random sampling from Gaussian distributions.

\begin{align} \label{eq:bayesian_rnn}
    \boldsymbol{\widetilde{W_t}} & = N(0,1)\times \log (1+e^{\rho_{\boldsymbol{W_t}}})+\mu_{\boldsymbol{W_t}} \\
    \boldsymbol{\widetilde{U_h}} & = N(0,1)\times \log (1+e^{\rho_{\boldsymbol{U_h}}})+\mu_{\boldsymbol{U_h}} \\
    \boldsymbol{\widetilde{b_h}} & = N(0,1)\times \log (1+e^{\rho_{\boldsymbol{b_h}}})+\mu_{\boldsymbol{b_h}} \\
    \boldsymbol{h_t} & = \sigma(\boldsymbol{\widetilde{W_t} \hat{h}_{t-1}} + \boldsymbol{\widetilde{U_h}}[ \boldsymbol{C_t}\circo \boldsymbol{m_t}] + \boldsymbol{\widetilde{b_h}})
\end{align}

\paragraph{Loss Function} Then Bayes by Backprop (BBB) \cite{graves2011practical} is used for training the BNNs, whose weights are sampled from a specific distribution. Here, $W^{i}$ denotes the $i^{th}$ Monte Carlo sample drawn from the variational posterior $Q(W^{i}\mid\theta)$.

\begin{align} \label{eq:bayesian_loss}
    \boldsymbol{\mathcal{L}(\theta)} & = -\mathbb{E}_{Q(\theta)}[\log P(\boldsymbol{X}\mid\theta)]+\boldsymbol{\text{KL}}[Q(\theta)\parallel P(\theta)] \\
&= \sum_{i=1}^{n}\{\log Q(W^{i}\mid\theta)-\log p(W^{i})-\log P(\boldsymbol{X}\mid W^{i})\}  \nonumber
\end{align}

Moreover, estimating confidence intervals does not come for free. The number of model parameters doubles due to each weight in the Bayesian layer being controlled by its distribution,  making the model easy to fall into local optimums.

Inspired by reinforcement learning, we design a training strategy to trade-off exploration (marginalization) and exploitation (optimization). The model either freezes the Bayesian layers to collapse to regular NN during training, or keep the open states. Freeze exploits the best model, while Unfreeze investigates uncertainty. The process alternates under preset conditions: Unfreeze every certain number of training steps.
\begin{table*}[t]
\centering
\resizebox{\linewidth}{!}{
\begin{tabular}{c|ccc|ccc|ccc}
\toprule
\multicolumn{1}{c}{} & \multicolumn{3}{c}{\textbf{Air}} & \multicolumn{3}{c}{\textbf{eICU}} & \multicolumn{3}{c}{\textbf{MIMIC-III (59)}} \\
\cmidrule{2-10}    \multicolumn{1}{c}{} & 5\%   & 10\%  & 20\%  & 5\%   & 10\%  & 20\%  & 5\%   & 10\%  & 20\% \\
    \midrule
    \textbf{Bayesian BRITS} & 0.0982 (0.1342) & 0.1022 (0.1401) & 0.1050 (0.1437) & 0.1690 (0.2191) & 0.1711 (0.2223) & 0.1755 (0.2279) & 0.09393 (0.23036) & 0.11176 (0.27545) & 0.13007 (0.32035) \\
    \textbf{Bayesian BRITS DML} & 0.0980 (0.1341) & \textbf{0.1001 (0.1372)*} & 0.1046 (0.1431) & 0.1693 (0.2194) & 0.1709 (0.2220) & 0.1750 (0.2273) & 0.09321 (0.22856) & 0.11230 (0.27677) & 0.13108 (0.32284) \\
    \textbf{Bayesian DEARI} & 0.0959 (0.1311) & 0.1012 (0.1386) & 0.1017 (0.1392) & \textbf{0.1582 (0.2051)*} & \textbf{0.1602 (0.2081)*} & \textbf{0.1645 (0.2136)*} & 0.09499 (0.23295) & 0.10904 (0.26876) & \textbf{0.12457 (0.30679)*} \\
    \textbf{Bayesian DEARI DML} & 0.0972 (0.1329) & 0.1006 (0.1379) & 0.1015 (0.1390) & 0.1584 (0.2052) & 0.1603 (0.2082) & 0.1646 (0.2138) & 0.09403 (0.23059) & 0.10981 (0.27065) & 0.12489 (0.30755) \\
    \textbf{BRITS DML} & 0.0996 (0.1362) & 0.1065 (0.1459) & 0.1076 (0.1473) & 0.1733 (0.2246) & 0.1704 (0.2213) & 0.1755 (0.2279) & 0.09532 (0.23380) & 0.11622 (0.28639) & 0.13151 (0.32389) \\
    \textbf{DEARI} & \textbf{0.0932 (0.1274)*} & 0.1089 (0.1492) & 0.1015 (0.1390) & 0.1608 (0.2085) & 0.1611 (0.2093) & 0.1654 (0.2148) & \textbf{0.09165 (0.22476)*} & \textbf{0.10789 (0.26591)*} & 0.12616 (0.31073) \\
    \textbf{DEARI DML} & 0.0934 (0.1277) & 0.1080 (0.1481) & \textbf{0.1014 (0.1387)*} & 0.1610 (0.2086) & 0.1615 (0.2098) & 0.1657 (0.2152) & 0.09168 (0.22485) & 0.10845 (0.26730) & 0.12602 (0.31036) \\
    \midrule
    \textbf{BRITS GRU} & 0.0996 (0.1362) & 0.1062 (0.1456) & 0.1095 (0.1499) & 0.1734 (0.2247) & 0.1703 (0.2212) & 0.1753 (0.2277) & 0.09581 (0.23501) & 0.11496 (0.28339) & 0.13109 (0.32289) \\
    \textbf{BRITS LSTM} & 0.1044 (0.1428) & 0.1203 (0.1649) & 0.1104 (0.1511) & 0.1676 (0.2172) & 0.1702 (0.2211) & 0.1757 (0.2282) & 0.09892 (0.24259) & 0.11453 (0.28226) & 0.13007 (0.32034) \\
    \textbf{V-RIN-full} & 0.1640 (0.2243) & 0.1658 (0.2273) & 0.1698 (0.2324) & 0.2308 (0.2992) & 0.2361 (0.3067) & 0.2481 (0.3222) & 0.10208 (0.25020) & 0.11713 (0.28865) & 0.13305 (0.32763) \\
    \textbf{GRUD}  & 5.8595 (8.0141) & 6.1388 (8.4140) & 6.7315 (9.2158) & 0.2213 (0.2868) & 0.2241 (0.2911) & 0.2293 (0.2977) & 0.39554 (0.97070) & 0.40830 (1.00721) & 0.44031 (1.08499) \\
    \textbf{MRNN}  & 0.2840 (0.3884) & 0.2919 (0.4000) & 0.3066 (0.4197) & 0.4602 (0.5964) & 0.4691 (0.6093) & 0.4863 (0.6316) & 0.25115 (0.61589) & 0.25336 (0.62415) & 0.25942 (0.63851) \\
    \midrule
    \multicolumn{1}{c}{} & \multicolumn{3}{c}{\textbf{MIMIC-III (89)}} & \multicolumn{3}{c}{\textbf{PhysioNet}} & \multicolumn{3}{c}{\textbf{Traffic}} \\
\cmidrule{2-10}    \multicolumn{1}{c}{} & 5\%   & 10\%  & 20\%  & 5\%   & 10\%  & 20\%  & 5\%   & 10\%  & 20\% \\
    \midrule
    \textbf{Bayesian BRITS} & 0.28065 (0.43634) & 0.29956 (0.46418) & 0.31491 (0.49188) & 0.2447 (0.3453) & 0.2521 (0.3556) & 0.2716 (0.3822) & 0.05967 (0.21919) & 0.07236 (0.26271) & 0.07951 (0.28776) \\
    \textbf{Bayesian BRITS DML} & 0.28103 (0.43693) & 0.30031 (0.46535) & 0.31540 (0.49264) & 0.2434 (0.3436) & 0.2518 (0.3551) & 0.2722 (0.3830) & 0.06039 (0.22187) & 0.07238 (0.26278) & 0.07952 (0.28783) \\
    \textbf{Bayesian DEARI} & 0.26684 (0.41487) & 0.28214 (0.43720) & 0.29590 (0.46218) & 0.2333 (0.3293) & 0.2415 (0.3406) & \textbf{0.2592 (0.3648)*} & 0.05515 (0.20262) & \textbf{0.06763 (0.24551)*} & \textbf{0.07423 (0.26865)*} \\
    \textbf{Bayesian DEARI DML} & 0.26629 (0.41401) & 0.28153 (0.43625) & 0.29571 (0.46188) & \textbf{0.2329 (0.3288)*} & 0.2417 (0.3409) & 0.2597 (0.3655) & 0.05503 (0.20219) & 0.06769 (0.24574) & 0.07425 (0.26871) \\
    \textbf{BRITS DML} & 0.28135 (0.43746) & 0.30088 (0.46625) & 0.31741 (0.49577) & 0.2484 (0.3506) & 0.2561 (0.3613) & 0.2766 (0.3892) & 0.05956 (0.21881) & 0.07265 (0.26374) & 0.08100 (0.29316) \\
    \textbf{DEARI} & \textbf{0.26496 (0.41196)*} & \textbf{0.28088 (0.43524)*} & 0.29535 (0.46132) & 0.2343 (0.3307) & 0.2412 (0.3402) & 0.2601 (0.3660) & \textbf{0.05479 (0.20128)*} & 0.06815 (0.24739) & 0.07530 (0.27254) \\
    \textbf{DEARI DML} & 0.26533 (0.41254) & 0.28091 (0.43530) & \textbf{0.29534 (0.46131)*} & 0.2337 (0.3298) & \textbf{0.2408 (0.3396)*} & 0.2599 (0.3658) & 0.05484 (0.20148) & 0.06851 (0.24868) & 0.07574 (0.27411) \\
    \midrule
    \textbf{BRITS GRU} & 0.28114 (0.43713) & 0.30092 (0.46631) & 0.31744 (0.49583) & 0.2477 (0.3496) & 0.2564 (0.3617) & 0.2772 (0.3901) & 0.05973 (0.21943) & 0.07295 (0.26486) & 0.08064 (0.29183) \\
    \textbf{BRITS LSTM} & 0.28406 (0.44164) & 0.30507 (0.47273) & 0.32253 (0.50379) & 0.2519 (0.3555) & 0.2629 (0.3709) & 0.2828 (0.3979) & 0.06000 (0.22046) & 0.07334 (0.26626) & 0.08074 (0.29221) \\
    \textbf{V-RIN-full} & 0.28530 (0.44357) & 0.30327 (0.46996) & 0.32234 (0.50348) & 0.2620 (0.3698) & 0.2734 (0.3857) & 0.2987 (0.4204) & 0.12967 (0.47662) & 0.13960 (0.50720) & 0.12831 (0.46444) \\
    \textbf{GRUD}  & 2.60073 (4.04391) & 2.65272 (4.11172) & 2.72067 (4.25052) & 0.4858 (0.6857) & 0.4960 (0.6998) & 0.5015 (0.7057) & 0.13054 (0.47970) & 0.13321 (0.48391) & 0.13789 (0.49914) \\
    \textbf{MRNN}  & 0.50953 (0.79225) & 0.51912 (0.80448) & 0.52620 (0.82191) & 0.5480 (0.7734) & 0.5550 (0.7830) & 0.5675 (0.7987) & 0.14788 (0.54355) & 0.15084 (0.54803) & 0.15468 (0.55995) \\
\bottomrule
\end{tabular}%
    }
\caption{The mean absolute error (MAE) and mean relative error (MRE) for all datasets. The best results are marked by *.}
\label{tab:1}%
\end{table*}%

\section{Experiments}

In this section, we evaluate and carefully analyse DEARI's performance against the state of the art using five real-world healthcare, environment, and traffic datasets. We compare DEARI against BRITS, GRUD, V-RIN(full) and MRNN. In all experiments, only the best model from each study is used for comparison. Although it would have been interesting to know how $E^2$GAN compares with other models, especially given that \cite{e2gan} does not provide a quantitative comparison with BRITS, we were unable to include $E^2$GAN in our study because the publicly-available code is based on the outdated TensorFlow 1.7 and Python 2.7, which are incompatible with our avaiflable GPU hardware.

\subsection{Datasets}
Each of the five datasets chosen for experimental evaluation has a different data distribution, especially the MIMIC-III database, from which we extracted two different datasets to model different tasks\footnote{The specific extraction method and statistical description of the data are detailed in the supplementary file.}. We reproduced the benchmarking papers for these public datasets, skipping steps that remove all-\textbf{NAN} samples to retain the data with its original missingness.

\paragraph{Air quality dataset.} Beijing Multi-Site Air-Quality Data \cite{zhang2017cautionary} provides hourly data on air pollutants from 12 locations. 17,532 samples with 18 variables.

\paragraph{eICU dataset.} The eICU Collaborative Research Database \cite{pollard2018eicu} is a public multi-center database with anonymized health data connected to more than 200,000 US ICU hospitalizations. 30,680 samples with 20 variables.

\paragraph{MIMIC-III dataset.} Medical Information Mart for Intensive Care III (MIMIC-III) \cite{johnson2016mimic}, an extensive, freely-available database of over 40,000 critical care patients based in Boston, Massachusetts between 2001 and 2012. We followed two benchmark papers and extracted different datasets for the experiment. 14,188 samples for 89 variables; 21,128 valid samples with 59 variables.

\paragraph{PhysioNet Challenge 2012 dataset.} The Predicting Mortality of ICU Patients: The PhysioNet/Computing in Cardiology Challenge 2012, a public medical benchmarking dataset provided by [Silva et al., 2012], contains records of 4000 48-hour ICU stays, allowing unbiased evaluations of different model performances. 3,997 samples with 35 variables.

\paragraph{Traffic dataset.} The Metro Interstate Traffic Volume Data Set is the public hourly volume of Interstate 94 Westbound traffic at MN DoT ATR station 301 on UCI Machine Learning Repository \cite{asuncion2007uci}. Hourly weather features and holidays included for impacts on traffic volume. 1,860 samples with 58 variables.





\subsection{Implementation Details}
We use the Adam optimizer and set the number of RNN hidden units to 108 for all models. The batch size is 64 for PhysioNet and traffic data and 128 for the rest. All datasets are normalised with zero mean and unit variance for stable training. We randomly chose 10\% from each dataset for validation and 10\% for testing, training on the remaining data. For the imputation task, we randomly mask $\{5\%, 10\%, 20\%\}$ observations in each dataset as the ground truth (validation data). 5-fold cross-validation is used to evaluate the models. We evaluate the imputation performance using mean absolute error (MAE) and mean relative error (MRE).

\subsection{Experimental Results}
Table \ref{tab:1} compares different combinations of DEARI components (a total of 7) and five baseline methods. The original BRITS uses an LSTM cell, but we also implemented a GRU-based BRITS, which performs better in 6 of our experiments, so we include it. The training process was made uniform across all experiments. To enable reproducible comparisons, we implement the 3-layer DEARI, using mean pooling and ['CLS'] token for DML on BRITS and DEARI respectively. Bayesian DEARI was implemented by unfreezing Bayesian layers every 100 steps and generating 10 simulations for all Bayesian components. The results show that DEARI outperforms BRITS in all experiments. The best DEARI model for the task, however, varies, and optimising the model for the task is part of our ongoing research. GRU BRITS still holds SOTA over baseline models, but almost all our components and their combinations outperform GRU BRITS in every dataset. Bayesian DEARI works better in large datasets.

\subsection{Ablation Study}
We now evaluate the different DEARI components using the PhysioNet Challenge 2012 dataset and a 10\% mask. 

\begin{figure*}[ht]
\centering
\subfigure[Increasing Depth]{
\begin{minipage}[t]{0.43\linewidth}{\label{fig:deari}}
\centering
\includegraphics[width=\linewidth,height=5cm]{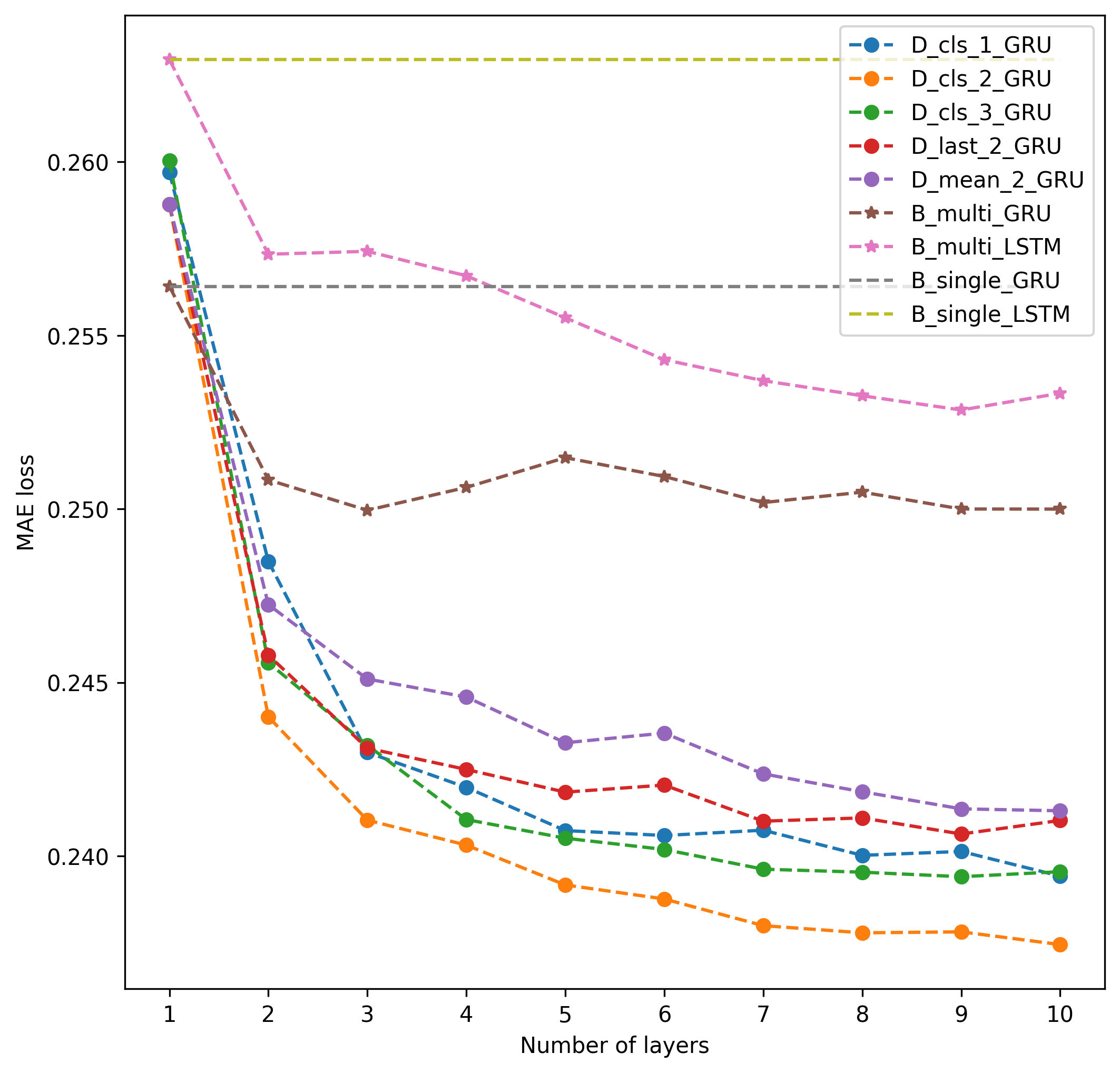}
\end{minipage}
}
\subfigure[Deep DML]{
\begin{minipage}[t]{0.43\linewidth}{\label{fig:deep_dml}}
\centering
\includegraphics[width=\linewidth,height=5cm]{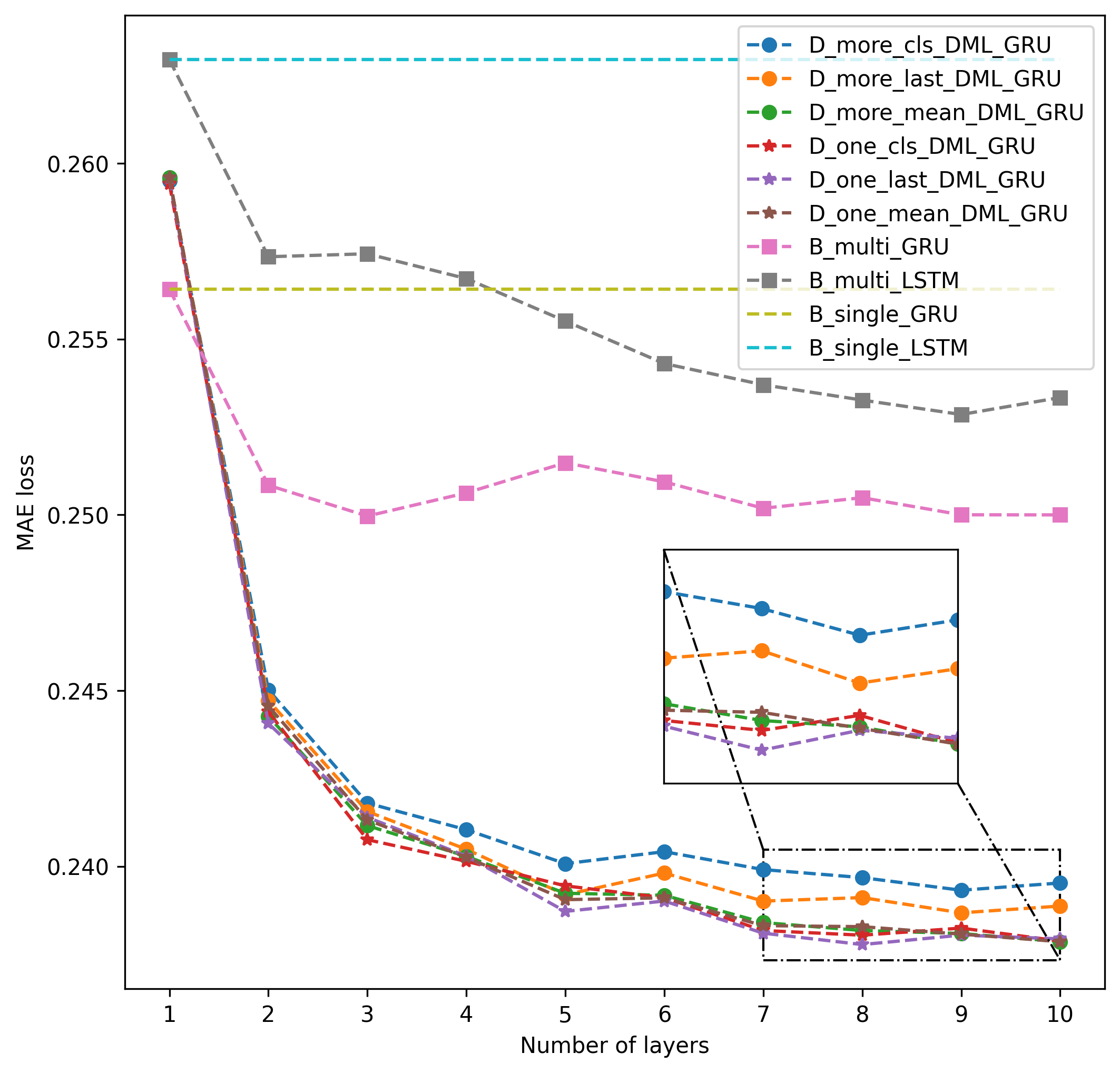}
\end{minipage}
}

 \subfigure[Confidence Interval for Systolic Blood Pressure]{
 \begin{minipage}[t]{0.43\linewidth}{\label{fig:ci}}
 \centering
 \includegraphics[width=\linewidth,height=5cm]{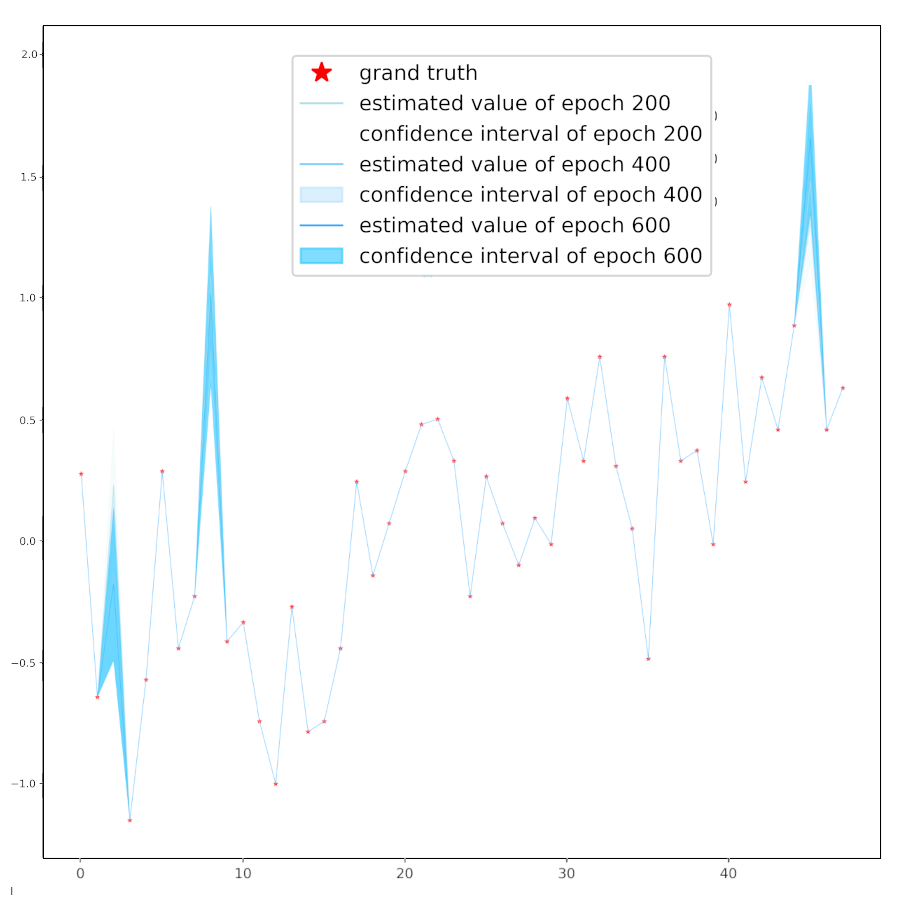}
 \end{minipage}
 }
 \subfigure[Model Complexity]{
\begin{minipage}[t]{0.43\linewidth}{\label{fig:complexity}}
\centering
\includegraphics[width=\linewidth,height=5cm]{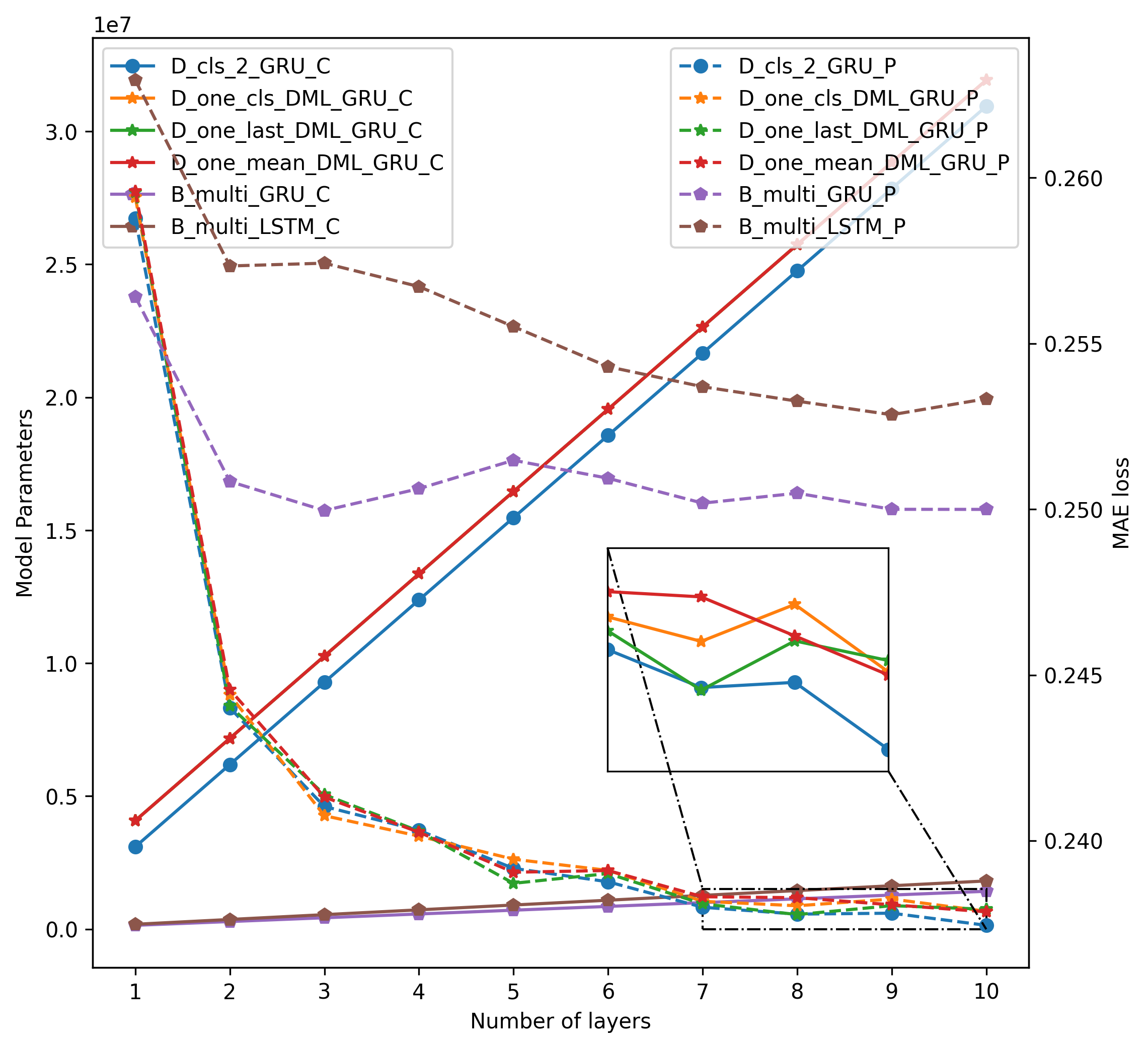}
\end{minipage}
}
\caption{Ablation Studies: \textbf{B:} BRITS; \textbf{D:} DEARI; \textbf{digit:} Number of encoder layers; \textbf{one/more:} Triple Size. In (d): \textbf{C}: Model complexity (number of parameters); \textbf{P}: Performance (MAE loss)}
\label{fig:3}
\end{figure*}

\paragraph{DEPTH} Fig \ref{fig:deari} shows the performance of different implementations of BRITS and DEARI when increasing the number of layers. Single-layer implementations of BRITS (B-single-GRU and B-single-LSTM) are used as baselines, holding constant MAE loss in the figure. The figure shows that the multilayer BRITS (B-multi-GRU and B-multi-LSTM) only slightly benefit from increasing model depth; MAE shows a modest decrease. This supports our observation that simple stacking without specific design is not a valid objective. For DEARI, although the benefit of increasing the number of layers slows down as models become deeper, the performance improvement is much more pronounced than in B-multi. Examining the different tokens, pooling and attention usage shows that the best performer is DEARI with a 2-self-attention encoder. The optimal performance is achieved by the 10-layer DEARI and CLS token representation, with MAE of 0.23745 (LSTM BRITS MAE = 0.2629 and GRU BRITS MAE = 0.2564). Please note that our reproduction of BRITS shows better performance than the original performance reported in \cite{cao2018brits} (MAE = 0.278). 

\paragraph{DML} Fig \ref{fig:deep_dml} summarises the outcomes of different DML strategies at varying depths. The figure not only demonstrates the superior ability of DEARI to benefit from DML, but also that self-attention can further improve the effectiveness of DML strategies. Moreover, increasing the number of triplets, which results in a bigger computational volume, degrades performance. It is worth noting that when DEARI is deep, the benefit from deep metric learning becomes negligible (although there is still a trend for further improvement). 

\paragraph{TRUSTWORTHY BOUNDARY} Figs \ref{fig:ci} shows an example of imputation confidence interval generated by the Bayesian model by sampling several well-optimised deterministic models. Confidence in the imputation can be quantified from corresponding boundaries; the more concentrated the result values, the smaller the confidence interval, signalling less uncertainty. As the figure shows, the estimation is generally successful in capturing the dynamic changes over time. The area of the shaded region is smaller when getting close to the observations, which is consistent with the temporal decay principle: the contribution of an observation to imputation decays with time.

\subsection{Model Complexity}
Model capability is directly related to the number of parameters. For DEARI, each additional 2-encoder layer increases the parameters by 2,149,330 (2$\times$90,577-parameter BRITS backbones, one in each direction, and 4$\times$492,044-parameter encoders, 2 in each direction). Fig \ref{fig:complexity} shows the relationship between model size and performance; the introduction of the transformer encoder (self attention) dominates both the parameter increase and the performance benefits. In other words, the proposed approach benefits mainly from the application of attention mechanisms. However, it is clear that model complexity and computational requirements grow linearly with the number of layers, while the performance improvement is eventually limited. This is the reason that we use 3-layer depth for all datasets. Nevertheless, compared to BRITS, DEARI's flexibility provides new possibilities for larger and more complex datasets. Reflecting on successful big transformer models, the BRITS backbone maybe a limiting factor constraining DEARI's generalization, which motivates us to consider pure attention models in future work.

\section{Discussion and Conclusions}
We propose a novel attention based deep recurrent neural network for heterogeneous time series imputation. Benefiting from awareness of uncertainty and heterogeneity, our DEARI surpasses state-of-the-art performance in real-world datasets, with explicit representation of imputation confidence.

In this work, we focused purely on imputation and have not examined downstream tasks (e.g., classification) because we regard pure imputation as a separate pre-processing module. Although some baseline models jointly train the imputer and predictor, they do not produce a fair comparative analysis of e.g. the effect of whether to pre-train or not, the effect of loss weights for the imputation and classification tasks and the effect of freezing the imputation layer after completing pre-training. Nevertheless, we conducted an initial investigation showing that MAE loss is comparable across classification outcomes (supplementary material). However, further exploration of the impact on downstream tasks requires more detailed exploration and discussion, and we hope to make it an extension of this work.

\bibliographystyle{named}
\bibliography{deari}

\end{document}